\documentclass[11pt]{article}

\usepackage[final]{acl}

\usepackage{times}
\usepackage{latexsym}
\usepackage[T1]{fontenc}
\usepackage[utf8]{inputenc}
\usepackage{microtype}
\usepackage{inconsolata}
\usepackage{graphicx}
\usepackage{booktabs}
\usepackage{xcolor}
\usepackage{multirow}
\usepackage{amsmath}
\usepackage[most]{tcolorbox}

\newtcolorbox{promptbox}[1][]{%
  colback=gray!5,
  colframe=gray!60,
  fonttitle=\bfseries\small,
  boxrule=0.5pt,
  arc=2pt,
  left=6pt, right=6pt, top=4pt, bottom=4pt,
  title={#1}
}

\newtcolorbox{examplebox}[1][]{%
  colback=white,
  colframe=gray!70,
  fonttitle=\bfseries\small,
  boxrule=0.5pt,
  arc=2pt,
  left=6pt, right=6pt, top=4pt, bottom=4pt,
  title={#1}
}

\newcommand{\down}[1]{{\color{red}\scriptsize$_{-#1}$}}

\title{Did You Forget What I Asked? \\Prospective Memory Failures in Large Language Models}

\author{Avni Mittal \\
  \texttt{avni.mittal2002@gmail.com}}

\begin{document}
\maketitle

\begin{abstract}
Large language models often fail to satisfy formatting instructions when they must simultaneously perform demanding tasks. We study this behavior through a \emph{prospective memory}-inspired lens from cognitive psychology, using a controlled paradigm that combines verifiable formatting constraints with benchmark tasks of increasing complexity. Across three model families and over 8,000 prompts, compliance drops by 2--21\% under concurrent task load. Vulnerability is highly type-dependent: terminal constraints (requiring action at the response boundary) degrade most, with drops up to 50\%, while avoidance constraints remain comparatively robust. A salience-enhanced format (explicit instruction framing plus a trailing reminder) recovers much of the lost compliance, restoring performance to 90--100\% in many settings. Interference is bidirectional: formatting constraints can also reduce task accuracy, with one model's GSM8K accuracy dropping from 93\% to 27\%. In additional stacking experiments, joint compliance declines sharply as constraints accumulate. All results use deterministic programmatic checkers, with no LLM-as-judge component, on publicly available datasets.
\end{abstract}

\section{Introduction}

Large language models are increasingly used in settings where users provide behavioral or formatting constraints alongside substantive requests. A typical prompt might ask for an explanation while also requiring a specific output form (e.g., all capitals, a required ending phrase, or a fixed number of bullets). Users expect the model to preserve such constraints throughout generation. In practice, models often produce strong content while violating the requested format.

We study this pattern as a \emph{functional analog} of prospective memory (PM): remembering to execute a deferred intention at the appropriate future point \cite{einstein2005prospective}. In humans, PM failures increase under cognitively demanding intervening tasks; this is often discussed as a cognitive-load effect. Our paper asks whether an analogous behavioral pattern appears in LLMs when instruction maintenance and task solving must be carried out together.

To test this, we compose verifiable IFEval-style constraints with benchmark tasks spanning low to high operational load (TriviaQA, MMLU, GSM8K, CNN/DailyMail). We evaluate both instruction compliance and task correctness with deterministic code-based checkers, avoiding LLM-as-judge components. Importantly, our load ordering is operational rather than a definitive cognitive scale, so we interpret load effects behaviorally.

Across three models from distinct architecture families and over 8,000 evaluated prompts, we find:

\begin{enumerate}
\item \textbf{Systematic forgetting under concurrent load.} Compliance drops 2--21\% when models must solve an additional task, with vulnerability varying sharply by instruction type (terminal constraints are most affected; avoidance constraints are nearly immune).
\item \textbf{A simple mitigation via salience.} A salience-enhanced prompt format recovers most lost compliance, restoring performance to 90--100\% in many settings.
\item \textbf{Dual-task tradeoffs and scaling fragility.} Interference is bidirectional (format constraints can reduce task accuracy, e.g., GSM8K 93\% to 27\%), and joint compliance declines as multiple constraints are stacked.
\end{enumerate}


\begin{figure*}[t]
\centering
\includegraphics[width=0.8\textwidth]{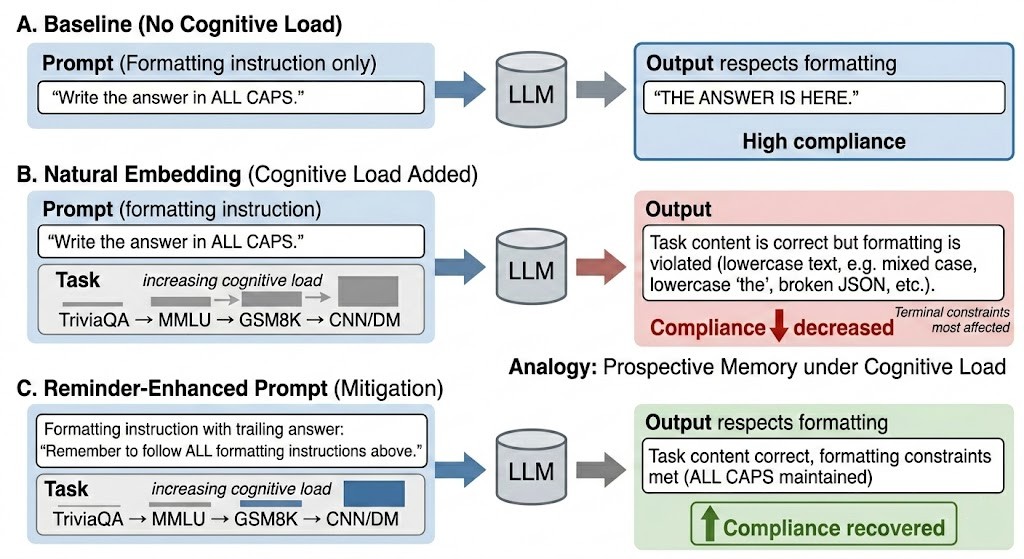}
\caption{Overview of the experimental pipeline. A verifiable formatting constraint from IFEval is composed with a benchmark task of varying difficulty (TriviaQA, MMLU, GSM8K, or CNN/DailyMail) using either the natural embedding template or the salience-enhanced reminder template. The model's response is then evaluated along two independent axes: (1) deterministic IFEval compliance checking (strict and loose) and (2) task-specific accuracy verification. This dual-evaluation design enables simultaneous measurement of prospective memory failure (compliance drop) and dual-task interference (accuracy drop).}
\label{fig:architecture}
\end{figure*}

\section{Related Work}

\paragraph{Instruction Following, Salience, and Robustness.} IFEval \cite{zhou2023instruction} established a practical, checker-based protocol for verifiable instruction compliance. More recent benchmarks broaden this picture: FollowBench shows that performance degrades as constraints are incrementally added \cite{jiang2024followbench}, while AGENTIF evaluates long, highly constrained agentic prompts \cite{qi2025agentif}. In parallel, work on instruction prioritization and robustness shows that failures can arise from conflict resolution or adversarial pressure rather than pure forgetting \cite{wallace2024instruction,zhang2025iheval,xiang2024certifiably}. We focus on a complementary failure mode: non-adversarial forgetting under concurrent task load.

\paragraph{Prospective Memory and Dual-Task Interference.} Our framing follows prospective-memory theory. Beyond the multiprocess account \cite{einstein2005prospective,mcdaniel2000strategic}, PAM-style evidence shows measurable ongoing-task costs when deferred intentions must be maintained \cite{smith2003cost}. Additional studies show that interference scales with intention characteristics and task complexity \cite{marsh2003interference}. This aligns with our finding that instruction type and concurrent-load level modulate both compliance and task accuracy. Methodologically, this perspective is consistent with the broader line of treating LLMs as subjects in cognitive experiments \cite{binz2023using,jones2022capturing}.

\paragraph{Long-Context and Retrieval Failures.} Prior long-context work, including Lost in the Middle \cite{liu2024lost}, documents strong position sensitivity in retrieval from long prompts. RULER further shows that long-context performance can degrade sharply on harder compositional retrieval tasks \cite{hsieh2024ruler}. Separately, GSM-IC demonstrates that irrelevant context can substantially harm reasoning accuracy \cite{shi2023large}. Our setting differs: the instruction is salient and fixed in position, but compliance still drops as concurrent demands rise, pointing to generation-time intention maintenance rather than pure retrieval failure.

\paragraph{Prompt Salience and Evaluation Reliability.} Prompt ordering and template choices materially affect model behavior, supporting reminder-based interventions \cite{chen2024premiseorder,guan2025ordereffect,lyu2024keepingaligned}. At evaluation time, LLMBar highlights weaknesses in judge-based assessment \cite{zeng2024llmbar}, reinforcing the value of deterministic, code-based checkers used in our experiments.


\section{Methodology}

Figure~\ref{fig:architecture} provides an overview of our experimental pipeline.

\subsection{Problem Formulation}

Given a prompt containing a formatting instruction $I$ and a task $T$, the model produces response $R$. We measure \emph{compliance rate} $CR$, defined as the fraction of responses passing the deterministic IFEval checker, and define the \emph{forgetting delta}:
\begin{equation}
\Delta = CR(I_{\text{alone}}) - CR(I + T)
\end{equation}

Positive $\Delta$ indicates the additional task caused the model to forget the instruction.

\subsection{Composition Framework}

\paragraph{Instructions (IFEval).} We use 15 of IFEval's 25 instruction types, selected for compatibility with task composition (the full list is given in Table~\ref{tab:types}, Appendix~\ref{app:details}). These span seven categories: case constraints, keyword requirements, terminal actions, structural formatting, avoidance rules, counting constraints, and length limits.

\paragraph{Distraction tasks.} We pair each IFEval constraint with a benchmark task at one of four cognitive load levels: TriviaQA (factual recall, low load), MMLU (multiple-choice reasoning, medium), GSM8K (multi-step arithmetic, high), and CNN/DailyMail (long-context summarization).

\subsection{Primary Experiment: Natural Embedding}
\label{sec:methodology_natural}

Each IFEval prompt already contains both a task and a constraint (e.g., ``Write an essay about Java. Use all capital letters.''). We append additional benchmark tasks using a neutral transition:

\begin{promptbox}[Natural Embedding Template]
\textit{\{original IFEval prompt with embedded constraint\}}

\vspace{4pt}
Additionally, please also complete the following:
\vspace{4pt}

\textit{\{benchmark question\}}
\end{promptbox}

The \emph{baseline} is the original IFEval prompt with no added task. The only variable between the baseline and the experimental conditions is the appended workload; the instruction format and salience are identical.

\subsection{Mitigation Experiment: Adding a Reminder}
\label{sec:methodology_reminder}

We test whether a simple prompt modification can recover compliance. We extract the constraint from IFEval metadata and present it with explicit framing and a trailing reminder sentence:

\begin{promptbox}[Reminder Template]
\textbf{IMPORTANT FORMATTING INSTRUCTION:}

\vspace{4pt}
\textit{\{extracted constraint text\}}

\vspace{4pt}
Now please help me with the following task:
\vspace{4pt}

\textit{\{benchmark question\}}

\vspace{4pt}
\textbf{Remember to follow ALL of my formatting instructions above.}
\end{promptbox}

This adds two modifications relative to the natural condition: (1) the ``IMPORTANT FORMATTING INSTRUCTION:'' prefix, and (2) the trailing reminder. Together, these test whether \emph{instruction salience}, i.e., making the constraint more prominent, can mitigate forgetting.

\subsection{Evaluation}

All evaluation is fully deterministic and uses code-based checkers with no LLM-as-judge component.

\paragraph{IFEval compliance.} Each of the 15 constraint types has a dedicated checker that returns binary pass/fail. We report two modes: \emph{strict} (applied to the raw model response) and \emph{loose} (the response passes if any of 8 normalized variants, e.g., stripping whitespace or markdown headers, satisfies the checker).

\paragraph{Task accuracy.} For GSM8K, we extract the last number from the response and compare it to the gold answer. For MMLU, we extract the answer letter using a priority heuristic and compare it to the gold label. For TriviaQA, we check whether any gold alias appears as a substring of the response (case-insensitive). For CNN/DailyMail, we compute ROUGE-L F1 with Porter stemming.

\begin{table*}[htbp]
\centering
\small
\setlength{\tabcolsep}{4.5pt}
\begin{tabular}{l ccc ccc}
\toprule
& \multicolumn{3}{c}{\textbf{Natural Embedding}} & \multicolumn{3}{c}{\textbf{With Reminder}} \\
\cmidrule(lr){2-4} \cmidrule(lr){5-7}
\textbf{Condition} & \textbf{o4-mini} & \textbf{DeepSeek} & \textbf{Llama} & \textbf{o4-mini} & \textbf{DeepSeek} & \textbf{Llama} \\
\midrule
Baseline & 86.1 & 89.3 & 88.9 & \multicolumn{3}{c}{--} \\
\midrule
+ TriviaQA & 82.2\down{3.9} & 83.0\down{6.3} & 87.8\down{1.1} & 93.3 & 92.2 & 92.2 \\
+ MMLU & 82.0\down{4.1} & 77.4\down{11.8} & 86.2\down{2.7} & 94.8 & 92.2 & 93.3 \\
+ GSM8K & 83.6\down{2.5} & 78.5\down{10.7} & 81.9\down{7.0} & 94.8 & 91.5 & 90.0 \\
+ 3$\times$GSM8K & 71.1\down{15.0} & 84.4\down{4.8} & 84.4\down{4.4} & 95.6 & 100.0 & 100.0 \\
+ 5$\times$GSM8K & 66.7\down{19.5} & 72.2\down{17.0} & 70.4\down{18.5} & 92.6 & 100.0 & 98.1 \\
+ CNN/DM & 77.5\down{8.6} & 78.9\down{10.4} & 86.8\down{2.1} & 98.4 & 100.0 & 100.0 \\
\bottomrule
\end{tabular}
\caption{IFEval strict compliance (\%) across all conditions. \textbf{Left}: natural embedding, where the formatting constraint is embedded in prose. Red subscripts show the forgetting delta vs.\ baseline. \textbf{Right}: salience-enhanced format with explicit framing and trailing reminder. The reminder recovers compliance to 90--100\% across all conditions, often exceeding the no-task baseline.}
\label{tab:main}
\end{table*}

\subsection{Multi-Constraint Stacking}
\label{sec:methodology_stacking}

Our primary experiments test one formatting constraint at a time. Real-world prompts, however, often contain multiple formatting requirements. To test whether forgetting compounds under stacking, we identify 11 of the 15 IFEval types that can be safely combined (excluding types with structural conflicts; see Table~\ref{tab:safe_pool}, Appendix~\ref{app:stacking_details}) and cross four factors: number of simultaneous constraints $N \in \{1, 2, 3, 5\}$, cognitive load $M \in \{0, 1, 3\}$ GSM8K problems, and trailing reminder $R \in \{\text{off}, \text{on}\}$. We additionally test 4 soft-tension constraint pairs to examine pair-specific interactions; the full design and prompt construction details are in Appendix~\ref{app:stacking_details}, with aggregate stacking results in Appendix~\ref{app:stacking_results}.

A response passes \emph{joint compliance} if it satisfies all $N$ constraint checkers simultaneously. We report both strict and loose variants.

\section{Experiments}

\subsection{Datasets and Models}

We compose prompts from five publicly available datasets (Table~\ref{tab:datasets}, Appendix~\ref{app:details}): IFEval \cite{zhou2023instruction} provides 90 stratified formatting constraints (6 per type $\times$ 15 types, fixed across runs); TriviaQA \cite{joshi2017triviaqa} supplies low-load factual recall; MMLU \cite{hendrycks2021mmlu} supplies medium-load multiple-choice reasoning; GSM8K \cite{cobbe2021gsm8k} supplies high-load multi-step arithmetic (tested in single, triple, and quintuple chains); and CNN/DailyMail \cite{see2017get} supplies long-context summarization. We evaluate three models spanning different architecture families: o4-mini (reasoning, Azure OpenAI), DeepSeek-V3.1 (open-weight, Azure AI), and Llama-3.3-70B-Instruct (instruction-tuned, Azure AI). All use greedy decoding and the same system prompt. Full dataset, model, and hyperparameter details are in Appendix~\ref{app:details}.

\begin{figure*}[t]
\centering
\includegraphics[width=\textwidth]{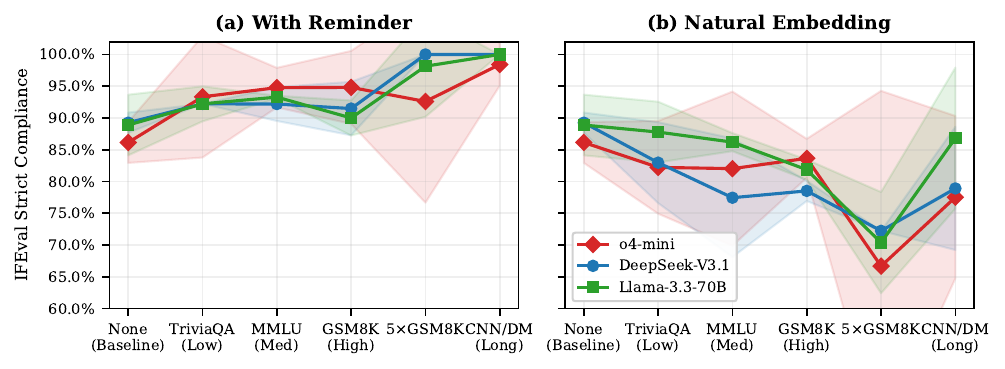}
\caption{Main result. IFEval compliance under increasing task complexity (a)~With salience-enhanced prompt: compliance stays flat at 90--100\%. (b)~Natural embedding: compliance drops consistently as distraction difficulty increases. Both conditions share the same baseline (no additional task).}
\label{fig:hero}
\end{figure*}

\subsection{Experimental Conditions}

Our experiments cross two factors: \emph{distraction task} (none, TriviaQA, MMLU, GSM8K, CNN/DailyMail) and \emph{prompt template} (natural embedding vs.\ reminder-enhanced). This yields the following conditions:

\paragraph{Baseline (IFEval-only).} The original IFEval prompt with no added task. This measures each model's intrinsic compliance rate and serves as the reference for computing forgetting deltas.

\paragraph{Task-only (no constraint).} Benchmark tasks presented without any formatting instruction. This provides a ceiling for task accuracy, against which we measure dual-task interference.

\paragraph{Natural embedding conditions.} For each of the four distraction tasks (TriviaQA, MMLU, GSM8K, CNN/DM), the benchmark question is appended to the original IFEval prompt with a neutral connector. Additionally, GSM8K is tested in chained configurations of 3 and 5 problems to intensify cognitive load, yielding six natural conditions in total.

\paragraph{Reminder-enhanced conditions.} The same six distraction conditions, but using the reminder template: the constraint is extracted and highlighted with ``IMPORTANT FORMATTING INSTRUCTION:'' at the top, and a trailing reminder sentence is appended at the bottom.

\subsection{Sampling, Evaluation, and Scale}

We conduct 3 independent runs (seeds 42, 137, 256) with the IFEval sample held fixed and benchmark items re-sampled per run. Within each run, all three models receive identical prompts. All evaluation is fully deterministic: IFEval compliance uses 15 code-based checkers (strict and loose modes); task accuracy uses last-number extraction (GSM8K), priority letter matching (MMLU), alias substring matching (TriviaQA), and ROUGE-L F1 (CNN/DM). Mean prompt and response token counts per condition are reported in Table~\ref{tab:prompt_length} (Appendix~\ref{app:lengths}). The full experiment comprises over 8,000 evaluated prompts. Detailed sampling, evaluation, and randomization procedures are in Appendix~\ref{app:protocols}.

\section{Results and Analysis}

\subsection{Forgetting and Recovery Under Load}

Table~\ref{tab:main} presents IFEval strict compliance under both experimental conditions. The left half (natural embedding) shows that appending additional tasks causes systematic compliance degradation across all three models. The right half (salience-enhanced reminder) shows that a simple prompt modification nearly eliminates this effect. All results report strict compliance; loose evaluation shows the same qualitative pattern with slightly attenuated deltas. Red subscripts (\down{x}) denote the forgetting delta relative to the baseline. Appendix~\ref{app:qualitative} presents a taxonomy of five forgetting mechanisms derived from manual analysis of all task-correct failures.

\begin{figure}[t]
\centering
\includegraphics[width=\columnwidth]{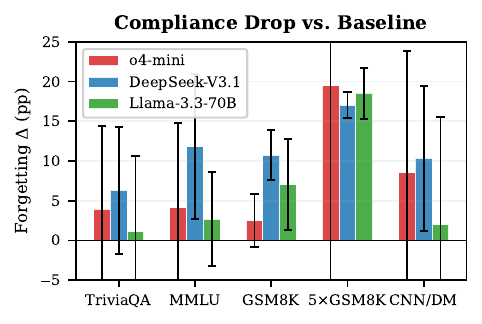}
\caption{Forgetting deltas by distraction type and model, with 95\% CI error bars from 3 runs. Positive values indicate compliance dropped vs.\ baseline. DeepSeek shows the largest forgetting; Llama is most robust.}
\label{fig:deltas}
\end{figure}

\paragraph{Forgetting.} Detailed statistical significance results are reported in Table~\ref{tab:stats} (Appendix~\ref{app:stats}). Table~\ref{tab:main} summarizes the overall forgetting trends. Compliance decreases under nearly every model--task combination, indicating broad-based forgetting. DeepSeek-V3.1 is the most affected (avg.\ $\Delta$ = 10.2\%), followed by o4-mini (avg.\ $\Delta$ = 8.9\%), while Llama-3.3-70B is the most robust (avg.\ $\Delta$ = 6.0\%). Figure~\ref{fig:hero}(b) visualizes the compliance curves, and Figure~\ref{fig:deltas} presents the per-condition compliance deltas with error bars.

\begin{figure}[htbp]
\centering
\includegraphics[width=\columnwidth]{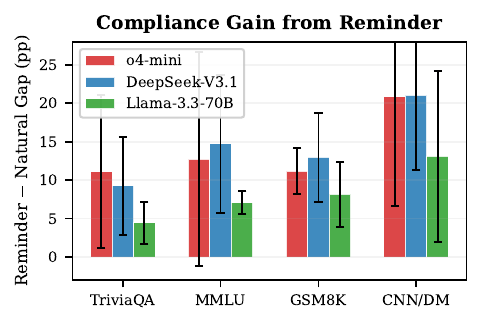}
\caption{Compliance gain from the salience-enhanced prompt format, by distraction type, with 95\% CI error bars. The effect is largest for the long-context condition (CNN/DailyMail (CNN/DM): +13 to +21\%).}
\label{fig:salience}
\end{figure}

\paragraph{Recovery.} The right half of Table~\ref{tab:main} shows that appending a single reminder sentence (``Remember to follow ALL of my formatting instructions above'') recovers compliance from the 67--87\% range to 90--100\% across all models and tasks. The recovery holds even under heavy chaining: at 3$\times$GSM8K, DeepSeek and Llama reach 100\% (vs.\ 84.4\% natural); at 5$\times$GSM8K, DeepSeek reaches 100\% (vs.\ 72.2\% natural) and Llama 98.1\% (vs.\ 70.4\%). The average compliance gain is 17.7\% for o4-mini, 16.9\% for DeepSeek, and 12.7\% for Llama (Figure~\ref{fig:salience}). The gain is largest for CNN/DM (+13 to +21\%), where the instruction is most temporally distant from the response end. Notably, the reminder prompts \emph{exceed} the no-task baseline in most cases.

\subsection{Terminal Constraints Are Most Vulnerable}
\label{sec:terminal}

Table~\ref{tab:pertype} decomposes forgetting by instruction type. The pattern is striking: \emph{terminal constraints}, those requiring action at the end of the response, suffer the largest drops. \texttt{end\_checker} drops 50\% for DeepSeek; \texttt{json\_format} drops 50\% for Llama; \texttt{number\_bullet\_lists} drops 44\% for DeepSeek. Meanwhile, avoidance constraints (\texttt{no\_comma}, \texttt{forbidden\_words}) are nearly immune to forgetting, holding at or near 100\%.

\begin{table}[t]
\centering
\small
\begin{tabular}{lccc}
\toprule
\textbf{Instruction Type} & \textbf{o4} & \textbf{DS} & \textbf{Llama} \\
\midrule
\multicolumn{4}{l}{\textit{Most vulnerable (baseline $\to$ +GSM8K, \%):}} \\
end\_checker & $-$17 & $-$50 & $-$17 \\
json\_format & +6 & $-$17 & $-$50 \\
number\_bullet\_lists & $-$12 & $-$44 & $-$11 \\
english\_capital & $-$17 & $-$17 & $-$22 \\
\midrule
\multicolumn{4}{l}{\textit{Most robust:}} \\
no\_comma & 0 & $-$11 & 0 \\
forbidden\_words & 0 & 0 & +6 \\
number\_words & 0 & 0 & $-$6 \\
\bottomrule
\end{tabular}
\caption{Per-type forgetting delta (baseline minus natural + GSM8K, in \%). We use the single GSM8K condition (rather than averaging across all tasks) because it represents the highest single-task cognitive load while keeping prompt structure uniform. Models shown are o4-mini, DeepSeek-V3.1, and Llama-3.3-70B-Instruct. Note: with only 6 items per type, individual deltas should be interpreted as indicative; the cross-type pattern is the robust finding.}
\label{tab:pertype}
\end{table}

This mirrors predictions from human prospective memory research. Constraints requiring \emph{continuous monitoring} (``never use a comma'') are maintained because every generated token provides an enforcement opportunity. Constraints requiring a \emph{deferred action} (``end with this exact phrase'') are forgotten because the relevant moment is temporally distant from the instruction. By the time the model reaches the end of its response, it has generated hundreds of tokens of task content, and the deferred instruction has lost effective salience.

This vulnerability pattern suggests a clear mitigation strategy: if forgetting occurs because the constraint loses salience during generation, then \emph{increasing salience} at the prompt boundary should recover compliance.

\subsection{Dual-Task Interference}

The forgetting effect is not one-directional. Adding formatting instructions also degrades performance on the benchmark task itself, revealing symmetric dual-task interference.

Table~\ref{tab:taskdrop} shows task accuracy under the reminder-enhanced condition compared against per-task baselines (no formatting constraint). MMLU accuracy drops 6--10\% across all models (e.g., DeepSeek: 77.8 $\to$ 68.4). The most dramatic interference appears in o4-mini's performance on chained GSM8K: accuracy on triplets drops from 93\% (single-GSM8K, task only) to just 27\% when a formatting constraint is simultaneously active. Llama and DeepSeek show smaller but consistent drops on chained math (61--78\% vs.\ the single-GSM8K ceiling of 88--94\%).

\begin{table}[t]
\centering
\small
\begin{tabular}{lccc}
\toprule
\textbf{Condition} & \textbf{o4-mini} & \textbf{DeepSeek} & \textbf{Llama} \\
\midrule
\multicolumn{4}{l}{\textit{Task-only baseline (no formatting constraint):}} \\
TriviaQA & 88.8 & 92.2 & 91.1 \\
MMLU & 82.2 & 77.8 & 81.1 \\
GSM8K (single) & 93.3 & 87.8 & 94.4 \\
\midrule
\multicolumn{4}{l}{\textit{With formatting constraint (reminder):}} \\
+ TriviaQA & 88.9 & 91.5 & 90.7 \\
+ MMLU & 76.5 & 68.4 & 71.3 \\
+ GSM8K$\times$1 & 85.9 & 84.4 & 83.3 \\
+ GSM8K$\times$3 & 26.7 & 76.7 & 64.4 \\
+ GSM8K$\times$5 & 53.7 & 77.8 & 61.1 \\
\midrule
+ CNN/DM$^\dagger$ & 18.5 & 20.2 & 19.4 \\
\bottomrule
\end{tabular}
\caption{Task accuracy (\%) with and without a formatting constraint. The task-only baseline averages accuracy across all single-task conditions (TriviaQA, MMLU, GSM8K$\times$1) without any formatting constraint. Adding IFEval instructions degrades benchmark performance, especially for o4-mini on chained GSM8K (93\% $\to$ 27\%). $^\dagger$CNN/DM reports ROUGE-L F1 ($\times$100); task-only baseline not available for this.}
\label{tab:taskdrop}
\end{table}

The o4-mini result is particularly striking. As a reasoning model, o4-mini appears to allocate its reasoning budget toward format compliance at the expense of mathematical accuracy, a tradeoff not observed as severely in the standard instruction-tuned models.

\begin{figure*}[t]
\centering
\includegraphics[width=\textwidth]{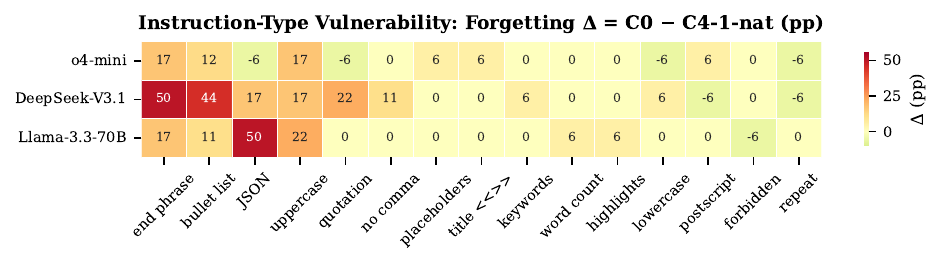}
\caption{Instruction-type vulnerability heatmap. Each cell shows the forgetting delta (baseline minus natural + GSM8K, in \%). Sorted by average vulnerability across models. Terminal and structural constraints cluster at the top (most forgotten); avoidance constraints cluster at the bottom (most robust).}
\label{fig:heatmap}
\end{figure*}

\begin{figure}[t]
\centering
\includegraphics[width=\columnwidth]{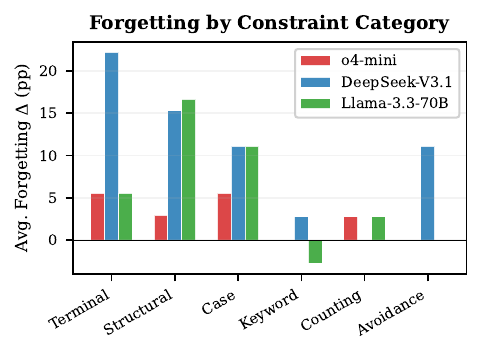}
\caption{Average forgetting delta by constraint category. Terminal and structural constraints show the largest drops across all models.}
\label{fig:categories}
\end{figure}

\begin{figure}[t]
\centering
\includegraphics[width=\columnwidth]{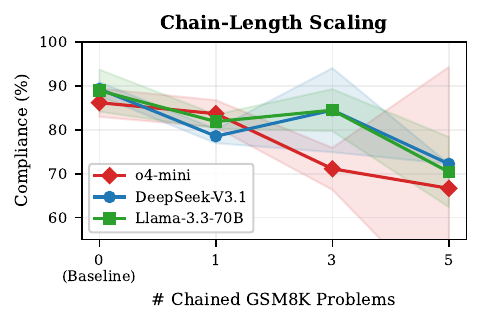}
\caption{Compliance vs.\ number of chained GSM8K problems.}
\label{fig:length}
\end{figure}

\paragraph{Non-monotonic scaling.} An unexpected pattern emerges in the chained GSM8K conditions (Figure~\ref{fig:length}): compliance does not decrease monotonically with chain length for two of the three models. DeepSeek's compliance peaks at GSM8K$\times$3 (84.4\%) before falling at GSM8K$\times$5 (72.2\%), exceeding even its single-GSM8K rate (78.5\%); Llama shows the same shape (81.9\% $\to$ 84.4\% $\to$ 70.4\%). For o4-mini, compliance decreases monotonically with chain length (83.6\% $\to$ 71.1\% $\to$ 66.7\%).

\subsection{Multi-Constraint Stacking Amplifies Forgetting}
\label{sec:stacking}

Our primary experiments test one constraint at a time. When multiple formatting constraints are stacked in a single prompt, the forgetting effect becomes much worse. Table~\ref{tab:stacking} shows joint compliance as a function of both constraint count ($N$) and cognitive load ($M$), averaged over the reminder condition ($R$). The full prompt construction is described in Appendix~\ref{app:stacking_details}; the complete $N \times M \times R$ grid, reminder effects, per-type vulnerability, and soft-tension analyses are in Appendices~\ref{app:stacking_results} and~\ref{app:stacking_tension}.

\begin{table*}[t]
\centering
\small
\setlength{\tabcolsep}{4pt}
\begin{tabular}{l ccc ccc ccc}
\toprule
& \multicolumn{3}{c}{\textbf{o4-mini}} & \multicolumn{3}{c}{\textbf{DeepSeek}} & \multicolumn{3}{c}{\textbf{Llama}} \\
\cmidrule(lr){2-4} \cmidrule(lr){5-7} \cmidrule(lr){8-10}
\textbf{N} & $M$=0 & $M$=1 & $M$=3 & $M$=0 & $M$=1 & $M$=3 & $M$=0 & $M$=1 & $M$=3 \\
\midrule
1 & 100.0 & 100.0 & 97.1 & 100.0 & 95.0 & 85.0 & 100.0 & 100.0 & 100.0 \\
2 & 94.9 & 97.5 & 85.0 & 85.0 & 70.0 & 70.0 & 100.0 & 100.0 & 85.0 \\
3 & 97.4 & 82.5 & 61.9 & 100.0 & 80.0 & 70.0 & 95.0 & 100.0 & 85.0 \\
5 & 55.0 & 55.0 & 32.5 & 65.0 & 55.0 & 65.0 & 95.0 & 80.0 & 75.0 \\
\bottomrule
\end{tabular}
\caption{Joint compliance (\%) under multi-constraint stacking, as a function of constraint count $N$ and cognitive load $M$ (number of concurrent GSM8K problems), averaged over reminder condition $R$. Reading left to right shows the effect of adding load; reading top to bottom shows the effect of adding constraints. Both factors compound: o4-mini at $N$=5, $M$=3 retains only 32.5\%.}
\label{tab:stacking}
\end{table*}

The table reveals two clear patterns. First, compliance decays steeply with $N$: o4-mini drops from near-100\% at $N$=1 to 32--55\% at $N$=5 depending on load, and DeepSeek drops to 55--65\%. Llama is most resilient, maintaining 75--95\% at $N$=5.

\paragraph{Cognitive load compounds stacking.} Reading left to right within each model panel, adding GSM8K problems alongside stacked constraints consistently degrades compliance. For o4-mini at $N$=5, compliance drops from 55.0\% ($M$=0) to 32.5\% ($M$=3). The same pattern holds across models: DeepSeek drops from 65.0\% to 65.0\% at $N$=5 (flat, but falls from 85.0\% to 70.0\% at $N$=2), and Llama drops from 95.0\% to 75.0\% at $N$=5. Stacking and cognitive load act as additive sources of forgetting.

\paragraph{The reminder becomes unreliable.} In the single-constraint experiments, the trailing reminder consistently recovers 10--17\% of lost compliance (Table~\ref{tab:main}). Under stacking, the picture changes. The matched reminder-on minus reminder-off differences are small on average: +5.8\% for o4-mini, +2.5\% for Llama, and +1.7\% for DeepSeek (Table~\ref{tab:stacking_reminder}). More importantly, the effect is inconsistent: it ranges from +25\% (o4-mini, $N$=3, $M$=1) to $-$20\% (o4-mini, $N$=2, $M$=3). In several high-load cells, adding the reminder actually \emph{hurts} compliance. A single generic reminder cannot maintain salience for multiple constraints the way it does for one.

\section{Discussion}
\label{sec:discussion}

\paragraph{Why do LLMs forget?} The formatting instruction remains fully visible in the context window throughout generation, so the forgetting we observe is not a retrieval failure in the traditional sense. We hypothesize it reflects \emph{representational competition}: when generating tokens for complex tasks, the model's hidden states become dominated by task-relevant representations, reducing the effective influence of the formatting constraint on token selection. This is analogous to the ``monitoring failure'' pathway in human PM, where attentional resources devoted to the ongoing task crowd out the monitoring process that would detect the PM cue. The reminder sentence re-elevates the constraint's attention weight at the critical final position, just before generation begins.

\paragraph{Connection to human prospective memory.} Our results replicate three classic human PM findings: (1) the \emph{cognitive load effect}, where PM drops with demanding ongoing tasks; (2) the \emph{cue salience effect}, where prominent cues improve PM (our reminder); and (3) the \emph{temporal distance effect}, where deferred actions are more forgotten than continuous ones (terminal vs.~pervasive constraints). The reminder works because it operates as a \emph{retrieval cue} placed at the highest-salience position in the prompt: immediately before generation, where recent tokens naturally receive high attention weight.

\paragraph{Salience as a design principle.} The reminder not only recovers lost compliance but often \emph{exceeds} the no-task baseline (Table~\ref{tab:main}), suggesting that standard instruction formats systematically underestimate model capability. Forgetting profiles also differ across architectures in ways that do not reduce to a simple load ordering. These patterns indicate that salience-enhancing interventions should be treated as a general prompt design principle rather than a remedial patch.

\paragraph{Practical implications.} For practitioners deploying LLMs: (1) always add a reminder sentence when combining formatting instructions with complex tasks; (2) be especially vigilant with terminal constraints; (3) test compliance under realistic cognitive load, not in isolation; (4) expect that adding formatting requirements will degrade task accuracy and budget for this tradeoff; (5) when stacking multiple formatting constraints, do not rely on a single generic reminder, since its effect becomes inconsistent under stacking; consider per-constraint reminders or breaking complex format requirements into separate prompts.

\section{Conclusion}

We introduced a composition-based paradigm inspired by prospective memory research to study instruction-following failures in LLMs. Across three models and over 8,000 prompts, we find that formatting compliance drops 2--21\% under concurrent task load, with terminal constraints most vulnerable; a simple salience-enhanced prompt recovers compliance to 90--100\%; and when multiple constraints are stacked, joint compliance falls below 50\% for one model and the reminder becomes unreliable. These findings suggest that instruction-tuning should reward compliance under load and that practitioners should treat terminal constraints as high-risk, apply reminders by default, and use structured approaches for multi-constraint scenarios.

\section{Limitations}

Our study has several limitations. We evaluate only three models from two cloud providers, limiting generalization to other architectures and scales. The IFEval sample provides only 6 items per instruction type, constraining per-type statistical power. The reminder condition modifies multiple prompt components simultaneously (constraint extraction, emphasis prefix, and trailing reminder), and we do not ablate their individual contributions. We test only single-turn scenarios; multi-turn settings may exhibit stronger forgetting. The cognitive load ordering is based on task complexity rather than empirically measured computational cost, and the non-monotonic patterns we observe suggest this ordering does not fully capture what drives forgetting. Finally, the stacking experiment uses a single run with 10 prompts per cell, limiting its statistical power.

\section*{Ethical considerations}

This work uses only publicly available datasets (IFEval, TriviaQA, MMLU, GSM8K, CNN/DailyMail) accessed through their standard distributions. No human subjects were involved. All experiments use greedy decoding (temperature~0) with fixed random seeds across three runs; note that exact bitwise reproducibility through cloud API endpoints is not guaranteed due to floating-point non-determinism, though generation-level outputs are expected to be stable. Our evaluation pipeline relies exclusively on code-based checkers with no LLM-as-judge component. Code and data will be released upon publication.

\bibliography{references}

\appendix
\section{Experimental Details}
\label{app:details}

\subsection{Instruction Types}

Table~\ref{tab:types} lists the 15 IFEval instruction types used in our experiments, grouped by category.

\begin{table}[h]
\centering
\small
\begin{tabular}{llc}
\toprule
\textbf{Category} & \textbf{Type} & \textbf{Count?} \\
\midrule
\multirow{2}{*}{Case} & english\_capital & No \\
& english\_lowercase & No \\
\midrule
\multirow{2}{*}{Keyword} & existence & No \\
& forbidden\_words & No \\
\midrule
\multirow{2}{*}{Terminal} & postscript & No \\
& end\_checker & No \\
\midrule
\multirow{3}{*}{Structural} & json\_format & No \\
& title & No \\
& quotation & No \\
\midrule
Avoidance & no\_comma & No \\
\midrule
\multirow{3}{*}{Counting} & num\_placeholders & Yes \\
& num\_bullet\_lists & Yes \\
& num\_highlighted & Yes \\
\midrule
Length & number\_words & Yes \\
\midrule
Sequential & repeat\_prompt & No \\
\bottomrule
\end{tabular}
\caption{15 IFEval instruction types used, grouped by category. ``Count?'' indicates whether the checker requires a numeric parameter.}
\label{tab:types}
\end{table}

\subsection{Models}

Table~\ref{tab:models} lists the three models used in our experiments. We deliberately select models from distinct architecture families to test whether prospective memory failures generalize beyond a single training paradigm. o4-mini is a reasoning-specialized model accessed through Azure OpenAI; DeepSeek-V3.1 is an open-weight model served via Azure AI Inference; and Llama-3.3-70B-Instruct is a 70-billion-parameter instruction-tuned model, also served through Azure AI. All three use the same system prompt and generation hyperparameters (\S4.2).

\begin{table}[h]
\centering
\small\setlength{\tabcolsep}{4pt}
\begin{tabular}{@{}llll@{}}
\toprule
\textbf{Model} & \textbf{Provider} & \textbf{Type} & \textbf{Params} \\
\midrule
o4-mini & Azure OpenAI & Reasoning & N/A \\
DeepSeek-V3.1 & Azure AI & Open-weight & N/A \\
Llama-3.3-70B & Azure AI & Instruct & 70B \\
\bottomrule
\end{tabular}
\caption{Models evaluated. All accessed via Azure-hosted API endpoints with greedy decoding (\texttt{temperature=0.0}, \texttt{max\_tokens=2048}).}
\label{tab:models}
\end{table}

\subsection{Datasets}

Table~\ref{tab:datasets} summarizes the five datasets used to construct our evaluation prompts. The IFEval sample provides the formatting constraints and is held fixed across all runs (seed~42) to ensure the same 90 items are tested in every condition. The four benchmark datasets serve as distraction tasks at increasing levels of cognitive demand: TriviaQA (single-step factual recall), MMLU (four-option reasoning), GSM8K (multi-step arithmetic), and CNN/DailyMail (long-context summarization). Benchmark items are re-sampled per run using distinct seeds (42, 137, 256) to provide variance estimates. GSM8K draws 270 items per run to support single, triplet, and quintuplet chain conditions. CNN/DailyMail articles are filtered to 600--1{,}000 tokens (measured with the \texttt{cl100k\_base} tokenizer) to ensure a consistent long-context condition.

\begin{table}[h]
\centering
\small
\begin{tabular}{llrrr}
\toprule
\textbf{Dataset} & \textbf{Split} & \textbf{Pool} & \textbf{Sampled} & \textbf{Load} \\
\midrule
IFEval & train & 541 & 90 & -- \\
TriviaQA & validation & 11,313 & 90/run & Low \\
MMLU & test & 14,042 & 90/run & Medium \\
GSM8K & test & 1,319 & 270/run & High \\
CNN/DM & test & -- & 45/run & Long \\
\bottomrule
\end{tabular}
\caption{Datasets and sampling. \textbf{Pool}: items in the source split. \textbf{Sampled}: items drawn per run. \textbf{Load}: cognitive load level assigned in our design. CNN/DM pool size varies after length filtering.}
\label{tab:datasets}
\end{table}

\section{Sampling, Evaluation, and Randomization Protocols}
\label{app:protocols}

\subsection{Statistical Details}
\label{app:stats}

We assess statistical significance using McNemar's test with continuity correction on paired binary outcomes (same IFEval items, baseline vs.\ condition) and report Cohen's~$h$ effect sizes ($h$ = 0.2: small, 0.5: medium, 0.8: large). Bootstrap 95\% confidence intervals are computed by resampling the 270 pooled item-level outcomes (90 items $\times$ 3 runs) with 10,000 bootstrap iterations. With 18 model--condition comparisons, individual $p$-values are not corrected for multiple testing; readers should interpret marginal results ($p \approx .05$) with appropriate caution.

Table~\ref{tab:stats} reports the full statistical results for the natural-embedding experiment.

\begin{table*}[h]
\centering
\small
\setlength{\tabcolsep}{4pt}
\begin{tabular}{ll cccc}
\toprule
\textbf{Model} & \textbf{Condition} & \textbf{Compliance [95\% CI]} & $\boldsymbol{\Delta}$ & \textbf{Cohen's} $\boldsymbol{h}$ & $\boldsymbol{p}$ \\
\midrule
\multirow{7}{*}{o4-mini} & Baseline & 86.1 [82.0, 90.3] & -- & -- & -- \\
& + TriviaQA & 82.2 [77.4, 86.7] & 3.9 & 0.11 & .063 \\
& + MMLU & 82.0 [77.2, 86.5] & 4.1 & 0.11 & .052 \\
& + GSM8K & 83.6 [79.2, 88.1] & 2.5 & 0.07 & .265 \\
& + 3$\times$GSM8K & 71.1 [61.1, 80.0] & 15.0 & 0.37 & .002\rlap{**} \\
& + 5$\times$GSM8K & 66.7 [53.7, 79.6] & 19.5 & 0.47 & .070 \\
& + CNN/DM & 77.5 [70.5, 84.5] & 8.6 & 0.23 & .010\rlap{**} \\
\midrule
\multirow{7}{*}{DeepSeek} & Baseline & 89.3 [85.6, 93.0] & -- & -- & -- \\
& + TriviaQA & 83.0 [78.5, 87.4] & 6.3 & 0.18 & $<$.001\rlap{***} \\
& + MMLU & 77.4 [72.2, 82.3] & 11.8 & 0.32 & $<$.001\rlap{***} \\
& + GSM8K & 78.5 [73.3, 83.3] & 10.7 & 0.30 & $<$.001\rlap{***} \\
& + 3$\times$GSM8K & 84.4 [76.7, 91.1] & 4.8 & 0.14 & .221 \\
& + 5$\times$GSM8K & 72.2 [59.3, 83.3] & 17.0 & 0.44 & .041\rlap{*} \\
& + CNN/DM & 78.9 [71.9, 85.9] & 10.4 & 0.29 & .006\rlap{**} \\
\midrule
\multirow{7}{*}{Llama} & Baseline & 88.9 [85.2, 92.6] & -- & -- & -- \\
& + TriviaQA & 87.8 [83.7, 91.5] & 1.1 & 0.04 & .546 \\
& + MMLU & 86.2 [82.1, 90.3] & 2.7 & 0.08 & .070 \\
& + GSM8K & 81.9 [77.0, 86.3] & 7.0 & 0.20 & $<$.001\rlap{***} \\
& + 3$\times$GSM8K & 84.4 [76.7, 91.1] & 4.4 & 0.13 & .013\rlap{*} \\
& + 5$\times$GSM8K & 70.4 [57.4, 81.5] & 18.5 & 0.47 & .004\rlap{**} \\
& + CNN/DM & 86.8 [80.6, 92.2] & 2.1 & 0.06 & .221 \\
\bottomrule
\end{tabular}
\caption{Statistical summary of forgetting under natural embedding. Compliance (\%) with bootstrap 95\% CIs (10k resamples). $\Delta$ = forgetting delta vs.\ baseline. Cohen's $h$: effect size (0.2 small, 0.5 medium). $p$: McNemar's test with continuity correction. Significance: {*}$p<.05$, {**}$p<.01$, {***}$p<.001$. Average $\Delta$ across conditions: DeepSeek 10.2\%, o4-mini 8.9\%, Llama 6.0\%.}
\label{tab:stats}
\end{table*}

\paragraph{Per-model analysis.}

\begin{itemize}

\item \textbf{DeepSeek-V3.1} shows the strongest forgetting (avg.\ $\Delta$ = 10.2\%). Its compliance drops significantly for all four single-task conditions ($p < .001$ for TriviaQA, MMLU, and GSM8K; $p = .006$ for CNN/DM). The largest single drop is on 5$\times$GSM8K ($\Delta$ = 17.0\%, $h$ = 0.44, a small-to-medium effect, $p = .041$).
\item \textbf{o4-mini} shows substantial forgetting (avg.\ $\Delta$ = 8.9\%). Single-task effects are marginal ($p = .052$--.063), but the drop reaches significance for chained GSM8K$\times$3 ($\Delta$ = 15.0\%, $h$ = 0.37, $p = .002$) and CNN/DM ($p = .010$), and is large in magnitude on 5$\times$GSM8K ($\Delta$ = 19.5\%, $h$ = 0.47, $p = .070$).
\item \textbf{Llama-3.3-70B} is the most robust on average (avg.\ $\Delta$ = 6.0\%), but shows significant degradation on +GSM8K ($\Delta$ = 7.0\%, $h$ = 0.20, $p < .001$), +3$\times$GSM8K ($p = .013$), and especially +5$\times$GSM8K ($\Delta$ = 18.5\%, $h$ = 0.47, $p = .004$). Most non-chained conditions show negligible effect sizes.
\end{itemize}

\subsection{Sampling and Randomization}

We conduct 3 independent runs using seeds 42, 137, and 256. The IFEval sample (90 items, stratified as 6 per type $\times$ 15 types) is drawn once with seed~42 and held fixed across all runs. The four benchmark datasets are re-sampled per run using distinct seeds so that cross-run variance reflects genuine sampling variability rather than item-specific effects.

Within each run, all three models receive the same set of prompts, enabling direct paired comparisons. Prompt ordering is deterministic within a run and consistent across models. GSM8K draws 270 items per run to support single, triplet, and quintuplet chain conditions. CNN/DailyMail articles are filtered to 600--1{,}000 tokens (measured with \texttt{cl100k\_base}) to ensure a consistent long-context condition.

\subsection{Evaluation Protocols}

All evaluation is fully deterministic and requires no LLM-as-judge.

\paragraph{IFEval compliance.} Each of the 15 constraint types has a dedicated code-based checker returning binary pass/fail. We report both \emph{strict} compliance (applied to the raw model response) and \emph{loose} compliance (applied to 8 normalized response variants: stripping leading/trailing whitespace, removing markdown headers, extracting content between quotation marks, etc.). A response passes loose compliance if \emph{any} variant passes.

\paragraph{Task accuracy.} GSM8K: we extract the last number from the response via regex and compare to the gold answer. MMLU: we extract the answer letter using a priority heuristic (``The answer is (X)'' $>$ first capital letter A--D $>$ regex fallback) and compare to the gold label. TriviaQA: we check whether any gold alias appears as a substring of the response (case-insensitive). CNN/DailyMail: we compute ROUGE-L F1 using the \texttt{rouge\_score} library with Porter stemming enabled; a score above 0.10 indicates task engagement.

\subsection{Inference Configuration}

All models use greedy decoding (\texttt{temperature=0.0}, \texttt{max\_tokens=2048}). For o4-mini, \texttt{temperature} is omitted (the API does not accept it for reasoning models) and \texttt{max\_completion\_tokens} replaces \texttt{max\_tokens}. All calls include a system prompt: ``You are a helpful assistant. Follow the user's instructions carefully and completely.'' API calls use exponential backoff with up to 5 retries; responses are checkpointed after every prompt to enable recovery from transient failures.

\section{Prompt and Response Length}
\label{app:lengths}

Table~\ref{tab:prompt_length} reports the mean input (prompt) and output (response) token counts per condition, averaged across all three models and all runs. The natural-embedding conditions have comparable prompt lengths to their reminder-enhanced counterparts, since the IFEval prompt text and benchmark content are similar in both templates. Response tokens are generally higher under natural embedding because the model also addresses the original IFEval creative task (e.g., writing an essay), whereas the reminder template extracts only the formatting constraint.

\begin{table}[h]
\centering
\small
\begin{tabular}{llrr}
\toprule
\textbf{Condition} & \textbf{Task} & \textbf{Input} & \textbf{Output} \\
\midrule
Baseline (C0) & -- & 72 & 538 \\
\midrule
\multicolumn{4}{l}{\emph{Reminder-enhanced}} \\
+ TriviaQA & Factual & 97 & 248 \\
+ MMLU & 4-choice & 166 & 308 \\
+ GSM8K$\times$1 & Arithmetic & 138 & 317 \\
+ GSM8K$\times$3 & 3$\times$Arith. & 279 & 644 \\
+ GSM8K$\times$5 & 5$\times$Arith. & 402 & 993 \\
+ CNN/DM & Summary & 891 & 324 \\
\midrule
\multicolumn{4}{l}{\emph{Natural embedding}} \\
+ TriviaQA & Factual & 98 & 558 \\
+ MMLU & 4-choice & 166 & 610 \\
+ GSM8K$\times$1 & Arithmetic & 139 & 624 \\
+ GSM8K$\times$3 & 3$\times$Arith. & 282 & 977 \\
+ GSM8K$\times$5 & 5$\times$Arith. & 393 & 1150 \\
+ CNN/DM & Summary & 892 & 629 \\
\bottomrule
\end{tabular}
\caption{Mean input and output token counts per condition, averaged across all models and runs. Token counts are from the API response metadata.}
\label{tab:prompt_length}
\end{table}

\section{Prompt Templates}
\label{app:prompts}

This section provides the exact prompt templates used in each experimental condition. All prompts are preceded by the system message: \texttt{``You are a helpful assistant. Follow all user instructions carefully.''}

\subsection{Baseline (C0): IFEval Only}

\begin{promptbox}[C0: Baseline]
\textit{\{original IFEval prompt, used verbatim\}}
\end{promptbox}

\noindent\textbf{Example:}

\begin{promptbox}[C0 Example: \texttt{english\_capital} + essay task]
\small
Write an article about how intra-team conflict affected sports teams. Write in a crazy coach screaming style. Use all capital letters to express the craziness. Basically, not a single word in your entire reply should contain lowercase letters.
\end{promptbox}

\subsection{Task-Only Baseline (C1)}

\begin{promptbox}[C1: Task Only (no formatting constraint)]
\textit{\{benchmark question only\}}
\end{promptbox}

\subsection{Natural Embedding (C2--C5-nat)}

\begin{promptbox}[Natural Embedding Template]
\small
\textit{\{original IFEval prompt with embedded constraint\}}

\vspace{4pt}
Additionally, please also complete the following:
\vspace{4pt}

\textit{\{benchmark question\}}
\end{promptbox}

\noindent\textbf{Example} (C2-nat: \texttt{english\_capital} + TriviaQA):

\begin{promptbox}[C2-nat Example]
\small
Write an article about how intra-team conflict affected sports teams. Write in a crazy coach screaming style. Use all capital letters to express the craziness. Basically, not a single word in your entire reply should contain lowercase letters.

\vspace{4pt}
Additionally, please also complete the following:
\vspace{4pt}

Which musical instrument was Jelly Roll Morton associated with?
\end{promptbox}

\subsection{Reminder-Enhanced (C2--C5)}

\begin{promptbox}[Reminder Template]
\small
IMPORTANT FORMATTING INSTRUCTION: \textit{\{extracted constraint text\}}

\vspace{4pt}
Now please help me with the following task:
\vspace{4pt}

\textit{\{benchmark question\}}

\vspace{4pt}
Please provide your response. Remember to follow ALL of my formatting instructions above.
\end{promptbox}

\noindent\textbf{Example} (C2: \texttt{english\_capital} + TriviaQA):

\begin{promptbox}[C2 Example]
\small
IMPORTANT FORMATTING INSTRUCTION: Your entire response should be in English, capital letters only.

\vspace{4pt}
Now please help me with the following task:
\vspace{4pt}

Which World War 2 American general was known as `Vinegar Joe'?

\vspace{4pt}
Please provide your response. Remember to follow ALL of my formatting instructions above.
\end{promptbox}

\subsection{Constraint Stacking}

\begin{promptbox}[Stacking Template (with reminder)]
\small
\textit{\{base topic\}} \textit{\{constraint 1\}} \textit{\{constraint 2\}} \ldots\ \textit{\{constraint N\}}

\vspace{4pt}
Also, please solve this math problem: \textit{\{GSM8K question\}}

\vspace{4pt}
Remember to follow all formatting instructions above.
\end{promptbox}

\noindent\textbf{Example} ($N$=3, $M$=1, reminder on):

\begin{promptbox}[Stacking Example]
\small
Write a short paragraph about renewable energy sources. The response must contain at least 8 placeholders represented by square brackets, such as [address]. Your answer must contain exactly 1 bullet points. Use the markdown bullet points such as: * This is a point. Your answer must contain a title, wrapped in double angular brackets, such as <<poem of joy>>.

\vspace{4pt}
Also, please solve this math problem: Stetson made a bet with Alec that he would give up \$10 for each orange he eats. While at the farm, Stetson ate 2/5 of the oranges they picked. If they picked 60 oranges, calculate the total amount of money Stetson gave up?

\vspace{4pt}
Remember to follow all formatting instructions above.
\end{promptbox}

\section{Additional Analysis Figures}
\label{app:additional_figures}

\subsection{Instruction Salience: Reminder vs.\ Natural Embedding}

\begin{figure*}[h]
\centering
\includegraphics[width=\textwidth]{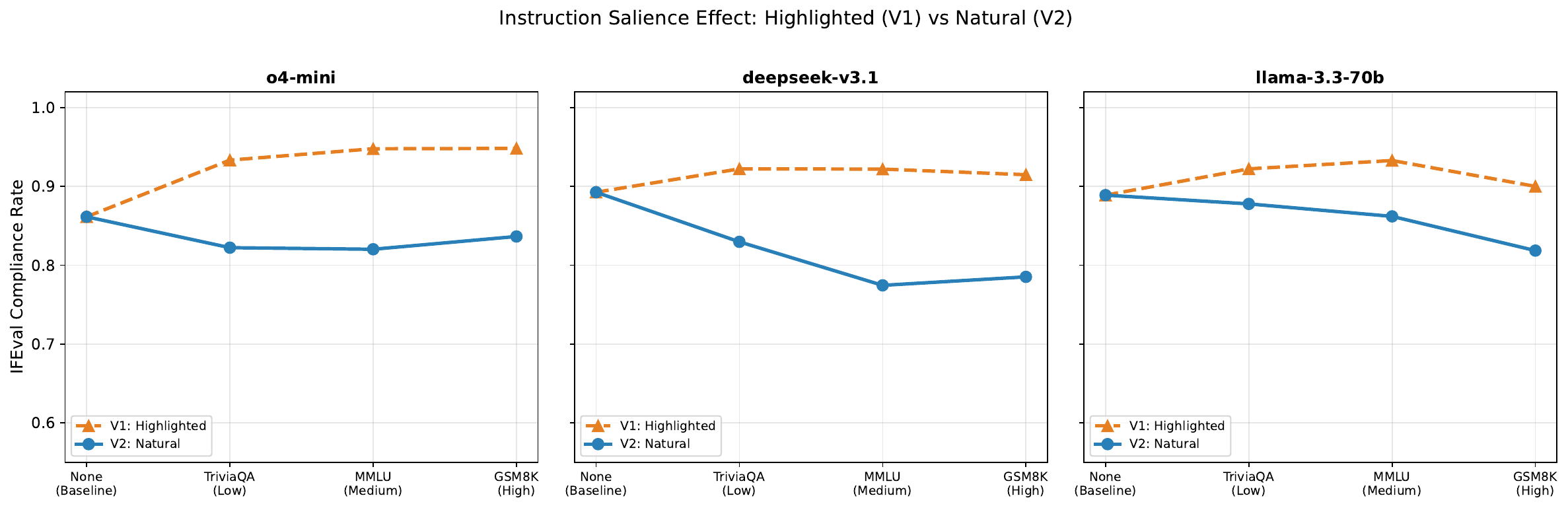}
\caption{Instruction salience effect. IFEval compliance under the reminder-enhanced format (V1, dashed) vs.\ natural embedding (V2, solid) across four distraction conditions, shown separately for each model. The gap between V1 and V2 reflects the compliance recovered by making the constraint more salient.}
\label{fig:v1_v2}
\end{figure*}

Figure~\ref{fig:v1_v2} compares IFEval compliance between the two prompt template designs across all four single-task conditions. The reminder-enhanced format (V1) consistently maintains higher compliance than the natural embedding (V2), with the gap widening as distraction difficulty increases. This visualizes the salience effect discussed in \S5.1 at the per-model level.

\subsection{Dual-Task Interference: Both Directions}

\begin{figure}[h]
\centering
\includegraphics[width=\columnwidth]{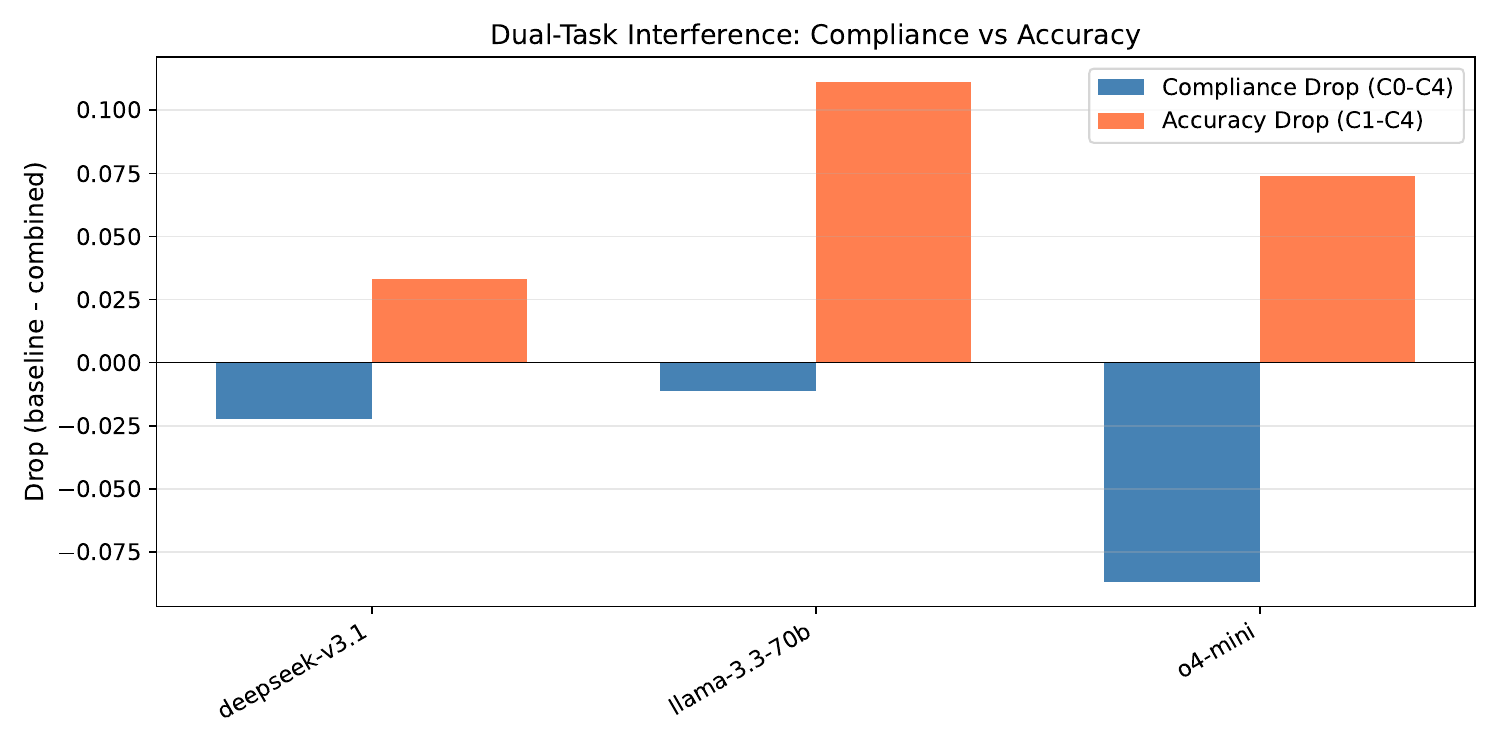}
\caption{Dual-task interference for each model under GSM8K. Blue bars: compliance drop (baseline IFEval compliance minus combined). Coral bars: accuracy drop (task-only GSM8K accuracy minus combined). Both directions show degradation, confirming symmetric interference.}
\label{fig:dual_task}
\end{figure}

Figure~\ref{fig:dual_task} visualizes the bidirectional cost of combining a formatting constraint with a benchmark task. The compliance drop (how much formatting adherence degrades) and the accuracy drop (how much task performance degrades) are shown side by side for each model under the GSM8K condition. This complements the task accuracy results in the main body by showing both interference directions in a single view.

\section{Constraint Stacking: Experimental Details}
\label{app:stacking_details}

\subsection{Safe Pool}

Table~\ref{tab:safe_pool} lists the 11 IFEval instruction types in the safe pool for the stacking experiment. Four types are excluded: \texttt{english\_lowercase} (directly conflicts with \texttt{english\_capital}), \texttt{json\_format} (forces a global structure incompatible with other formatting), \texttt{end\_checker} (overlaps with \texttt{postscript} and \texttt{quotation}), and \texttt{repeat\_prompt} (prepends a verbatim copy that interferes with other constraints).

\begin{table}[h]
\centering
\small
\resizebox{\columnwidth}{!}{%
\begin{tabular}{ll}
\toprule
\textbf{Type} & \textbf{Description} \\
\midrule
english\_capital & Entire response in capital letters \\
existence & Include specified keywords \\
forbidden\_words & Avoid specified keywords \\
postscript & End with a postscript (P.S.) \\
title & Include a title in angular brackets \\
no\_comma & Avoid all commas \\
number\_placeholders & Include $N$ placeholders in brackets \\
number\_bullet\_lists & Include exactly $N$ bullet points \\
number\_words & Response within $N$ words \\
number\_highlighted & Highlight $N$ sections with markdown \\
quotation & Wrap response in double quotes \\
\bottomrule
\end{tabular}}
\caption{The 11 IFEval instruction types in the safe pool for constraint stacking.}
\label{tab:safe_pool}
\end{table}

Two hard-conflict pairs are prevented from co-occurring within a prompt:

\begin{itemize}
\item \texttt{postscript}~+~\texttt{quotation}: both impose boundary requirements.
\item \texttt{no\_comma}~+~\texttt{number\_bullet\_lists}: bullet formatting conventionally requires commas.
\end{itemize}

\subsection{Soft-Tension Pairs}

Four pairs are tested separately to examine pair-specific interactions:

\begin{enumerate}
\item \texttt{no\_comma} + \texttt{number\_bullet\_lists}: Bullet lists conventionally use commas, making comma avoidance harder.
\item \texttt{postscript} + \texttt{quotation}: Both constrain response boundaries (postscript at the end, quotation marks wrapping everything).
\item \texttt{english\_capital} + \texttt{existence}: Required keywords may conflict with capitalization conventions for domain terms.
\item \texttt{number\_words} + \texttt{number\_bullet\_lists}: Hitting a word count target while producing exactly $N$ bullets can create competing pressures.
\end{enumerate}

\subsection{Prompt Construction and Grid}

\begin{table*}[h]
\centering
\small
\setlength{\tabcolsep}{3.5pt}
\begin{tabular}{l rr rr rr rr rr rr}
\toprule
& \multicolumn{4}{c}{\textbf{o4-mini}} & \multicolumn{4}{c}{\textbf{DeepSeek}} & \multicolumn{4}{c}{\textbf{Llama}} \\
\cmidrule(lr){2-5} \cmidrule(lr){6-9} \cmidrule(lr){10-13}
& \multicolumn{2}{c}{$M$=0} & \multicolumn{2}{c}{$M$=3} & \multicolumn{2}{c}{$M$=0} & \multicolumn{2}{c}{$M$=3} & \multicolumn{2}{c}{$M$=0} & \multicolumn{2}{c}{$M$=3} \\
\cmidrule(lr){2-3} \cmidrule(lr){4-5} \cmidrule(lr){6-7} \cmidrule(lr){8-9} \cmidrule(lr){10-11} \cmidrule(lr){12-13}
$N$ & off & on & off & on & off & on & off & on & off & on & off & on \\
\midrule
1 & 100 & 100 & 94 & 100 & 100 & 100 & 80 & 90 & 100 & 100 & 100 & 100 \\
2 & 95 & 95 & 95 & 75 & 80 & 90 & 80 & 60 & 100 & 100 & 90 & 80 \\
3 & 100 & 95 & 50 & 74 & 100 & 100 & 80 & 60 & 90 & 100 & 80 & 90 \\
5 & 50 & 60 & 25 & 40 & 60 & 70 & 50 & 80 & 100 & 90 & 70 & 80 \\
\bottomrule
\end{tabular}
\caption{Full stacking grid: joint compliance (\%) by $N$, $M$ ($M$=1 omitted for space; pattern interpolates), and $R$ (off/on). Cells where the reminder hurts: o4-mini $N$=2/$M$=3 (75 vs.\ 95, $\Delta$=$-$20), DeepSeek $N$=5/$M$=1 (40 vs.\ 70, $\Delta$=$-$30). Cells where it helps most: o4-mini $N$=3/$M$=3 (74 vs.\ 50, $\Delta$=+24).}
\label{tab:stacking_full}
\end{table*}

Each stacking prompt is assembled from four components: (1)~a neutral base topic (one of 15 subjects such as ``the history of bridges'' or ``how weather forecasting works''), (2)~$N$ constraint instructions with realistic parameters sampled from IFEval metadata, (3)~$M$ GSM8K problems drawn from the test set, and (4)~an optional trailing reminder (``Remember to follow ALL of the formatting instructions listed above.''). Constraints are listed as labeled items (Constraint 1, Constraint 2, etc.) before the base topic, and GSM8K problems are appended after the topic.

The safe-pool grid crosses $N \in \{1, 2, 3, 5\}$, $M \in \{0, 1, 3\}$, and $R \in \{\text{off}, \text{on}\}$ (24 cells, 10 prompts each, 240 total). The tension grid fixes $N$=2 and crosses 4 pairs $\times$ $M \times R$ (24 cells, 10 prompts each, 240 total). Combined: 480 prompts $\times$ 3 models = 1,440 API calls.

\section{Stacking: Full Results}
\label{app:stacking_results}

\subsection{Full Stacking Grid ($N \times M \times R$)}

Table~\ref{tab:stacking_full} reports joint compliance for every cell of the stacking design, broken down by constraint count ($N$), cognitive load ($M$), and reminder ($R$). This backs up the cell-level claims in \S\ref{sec:stacking}: the reminder lift ranges from +25\% (o4-mini, $N$=3, $M$=1) to $-$20\% (o4-mini, $N$=2, $M$=3).

\subsection{Reminder Effect Under Stacking}

Let $J_m(N,M,R)$ denote joint compliance for model $m$ at a fixed number of constraints $N$, cognitive load $M$, and reminder setting $R$. The reminder effect reported in Table~\ref{tab:stacking} is the matched on--off difference averaged over the 12 safe-pool cells:
\begin{align}
\Delta_R^{(m)} &= \frac{1}{12}\sum_{N \in \{1,2,3,5\}} \sum_{M \in \{0,1,3\}} d_m(N,M), \\
d_m(N,M) &= J_m(N,M,\mathrm{on}) - J_m(N,M,\mathrm{off}).
\end{align}

\begin{table}[h]
\centering
\small
\begin{tabular}{lccc}
\toprule
\textbf{Model} & \textbf{$R$=off} & \textbf{$R$=on} & \textbf{$\Delta_R$} \\
\midrule
o4-mini & 77.0 & 82.8 & +5.8 \\
DeepSeek & 77.5 & 79.2 & +1.7 \\
Llama & 91.7 & 94.2 & +2.5 \\
\bottomrule
\end{tabular}
\caption{Mean joint compliance (\%) by reminder setting under stacking. Values average over all 12 safe-pool $N \times M$ cells. $\Delta_R$ is the final column minus the middle column and corresponds to the $\Delta_R$ column in Table~\ref{tab:stacking}.}
\label{tab:stacking_reminder}
\end{table}

\subsection{Per-Type Vulnerability Under Stacking}

The vulnerability pattern from the single-constraint experiments carries over under stacking. Avoidance constraints (\texttt{no\_comma}: 100\% pass rate across all models, \texttt{forbidden\_words}: 99--100\%) remain nearly immune even at $N$=5. Counting and content-insertion constraints are most vulnerable: \texttt{number\_highlighted\_sections} averages only 71--77\% for o4-mini and DeepSeek. Table~\ref{tab:stacking_pertype} shows the full per-type breakdown.

\begin{table}[h]
\centering
\small
\resizebox{\columnwidth}{!}{%
\begin{tabular}{lccc}
\toprule
\textbf{Constraint Type} & \textbf{o4-mini} & \textbf{DeepSeek} & \textbf{Llama} \\
\midrule
num\_highlighted\_sections & 76.6 & 71.4 & 96.4 \\
number\_bullet\_lists & 81.9 & 75.0 & 97.2 \\
quotation & 77.2 & 91.7 & 100.0 \\
number\_placeholders & 82.2 & 88.4 & 100.0 \\
number\_words & 88.3 & 87.0 & 95.9 \\
existence & 76.7 & 95.6 & 100.0 \\
postscript & 77.8 & 100.0 & 98.6 \\
english\_capital & 97.5 & 97.5 & 90.4 \\
title & 86.2 & 100.0 & 100.0 \\
forbidden\_words & 99.0 & 100.0 & 100.0 \\
no\_comma & 100.0 & 100.0 & 100.0 \\
\bottomrule
\end{tabular}}
\caption{Per-type pass rate (\%) under stacking, sorted by average vulnerability across models. Counting constraints are most vulnerable; avoidance constraints remain immune.}
\label{tab:stacking_pertype}
\end{table}

\section{Stacking: Soft-Tension Ablation}
\label{app:stacking_tension}

Table~\ref{tab:tension_summary} shows the average joint compliance gap between each tension pair and the safe-pool $N$=2 baseline, averaged across all $M$ and $R$ settings. A negative value means the tension pair performs worse than the safe-pool average at $N$=2.

\begin{table}[h]
\centering
\small
\begin{tabular}{lccc}
\toprule
\textbf{Tension Pair} & \textbf{o4-mini} & \textbf{DS} & \textbf{Llama} \\
\midrule
no\_comma + bullet\_lists & $-$17.5 & +21.7 & $-$6.7 \\
postscript + quotation & +5.9 & +23.3 & +1.7 \\
english\_cap + existence & $-$1.9 & +16.7 & 0.0 \\
num\_words + bullet\_lists & $-$21.6 & $-$16.7 & $-$15.0 \\
\bottomrule
\end{tabular}
\caption{Average tension penalty ($\Delta$ in \%, tension minus safe $N$=2). DS = DeepSeek-V3.1. Negative means the tension pair degrades compliance below the safe-pool baseline. The \texttt{number\_words}~+~\texttt{number\_bullet\_lists} pair shows the largest penalty across all models.}
\label{tab:tension_summary}
\end{table}

The \texttt{number\_words}~+~\texttt{number\_bullet\_lists} pair consistently degrades compliance across all three models ($-$15\% to $-$22\% on average), confirming that meeting a word count target while also producing a specific number of bullets creates real interference. The \texttt{postscript}~+~\texttt{quotation} pair, despite being excluded from the safe pool as a hard conflict, performs at or above the safe-pool baseline. This suggests that its classification as a hard conflict was conservative; the two constraints may actually be complementary rather than competing, since both are boundary-focused requirements that the model can satisfy with a single response structure.

Note that the safe-pool $N$=2 baseline uses randomly drawn constraint pairs and includes all 11 types, some of which (e.g., \texttt{number\_highlighted\_sections}) have low individual pass rates. Tension pairs composed of individually easy constraints (like \texttt{no\_comma} or \texttt{postscript}) may therefore appear to outperform the safe-pool average even if they impose real tension, because their constituent types are inherently easier. The most meaningful comparison is for pairs where both constraints have moderate individual pass rates, making the \texttt{number\_words}~+~\texttt{number\_bullet\_lists} result (consistently negative across all models) the cleanest signal of pair-specific interference.

\section{Taxonomy of Forgetting Mechanisms}
\label{app:qualitative}

To understand \emph{how} models forget, we manually inspected all 187 instances in which a model completed the distractor task correctly\footnote{For CNN/DailyMail, where task correctness is a continuous metric, we use ROUGE-L~$>$~0.10 as the criterion.} but violated the formatting constraint it satisfied in the C0 baseline. We identified five recurring failure mechanisms, summarized in Table~\ref{tab:mechanisms}; one representative example for each follows. Examples are drawn from different models, conditions, and distractor tasks to illustrate the generality of the phenomenon.

\begin{table}[h]
\centering
\small
\begin{tabular}{lrp{3.7cm}}
\toprule
\textbf{Mechanism} & $N$ & \textbf{Description} \\
\midrule
Domain shift & 55 & Constraint held for IFEval subtask, dropped at task transition \\
Structural abandon & 51 & Formatting maintained for first subtask, replaced for second \\
Terminal omission & 44 & Required ending produced mid-response; generation continues \\
Spatial compartm. & 19 & Global constraint applied only to one output section \\
Content/length & 18 & Required elements or length constraints not met \\
\bottomrule
\end{tabular}
\caption{Taxonomy of forgetting mechanisms. $N$ = number of task-correct forgetting failures across all models and runs (187 total).}
\label{tab:mechanisms}
\end{table}

\subsection{Domain Shift: Case Violation Under TriviaQA}
\label{app:ex_uppercase}

Llama-3.3-70B, reminder-enhanced condition with TriviaQA (C2). The constraint requires the entire response to be in capital letters.

\paragraph{Analysis.} The model writes 95.1\% of alphabetic characters in uppercase but reverts to title case for the proper noun ``Binnacle'' (14 of 288 alphabetic characters are lowercase). The model's world knowledge about conventional capitalization of domain-specific terms overrides the active formatting constraint. The task answer is correct, and the constraint would have been satisfied in the C0 baseline.

\begin{examplebox}[Example: Domain Shift (Case)]
\small
\colorbox{blue!6}{\parbox{\dimexpr\linewidth-2\fboxsep}{%
\textbf{Prompt}\\[2pt]
\texttt{IMPORTANT FORMATTING INSTRUCTION: Your entire response should be in English, capital letters only.}\\
\texttt{Now please help me with the following task:}\\
\texttt{What is the (traditionally wooden) housing or stand which contains a ship's compass?}\\
\texttt{Please provide your response. Remember to follow ALL of my formatting instructions above.}
}}

\vspace{4pt}
\colorbox{red!6}{\parbox{\dimexpr\linewidth-2\fboxsep}{%
\textbf{Response}\\[2pt]
\texttt{THE HOUSING OR STAND WHICH CONTAINS A SHIP'S COMPASS IS CALLED A BIMINI. HOWEVER, THE MORE TRADITIONAL TERM FOR THIS IS A \textbf{Binnacle}. A BIMINACLE IS ALSO SOMETIMES USED TO REFER TO THIS, BUT \textbf{Binnacle} IS THE MORE COMMONLY USED TERM. IT IS TYPICALLY A WOODEN HOUSING THAT CONTAINS AND PROTECTS THE COMPASS, WHILE ALSO PROVIDING A STABLE PLATFORM FOR IT TO SIT ON.}
}}
\end{examplebox}


\subsection{Domain Shift: Punctuation Forgotten During Arithmetic}
\label{app:ex_comma}

DeepSeek-V3.1, natural-embedding condition with GSM8K (C4-1-nat). The constraint is to avoid all commas.

\begin{examplebox}[Example: Domain Shift (Punctuation)]
\small
\colorbox{blue!6}{\parbox{\dimexpr\linewidth-2\fboxsep}{%
\textbf{Prompt}\\[2pt]
\texttt{Rewrite the sentence ``A bust of a man with a beard and mustache.'' in a more sophisticated way. Do not use commas in your response.}\\
\texttt{Additionally, please also complete the following:}\\
\texttt{Zeke's baseball team has 7 more players than Carlton's. If Carlton's team has 13 players, how many players are there in both teams combined?}
}}

\vspace{4pt}
\colorbox{red!6}{\parbox{\dimexpr\linewidth-2\fboxsep}{%
\textbf{Response}\\[2pt]
\texttt{A sculpted bust depicts a bearded and mustachioed gentleman.}\\
\texttt{Carlton's team has 13 players. Zeke's team has 7 more\textbf{,} so it has 13 + 7 = 20 players. Combined\textbf{,} both teams have 13 + 20 = 33 players.}
}}
\end{examplebox}

\paragraph{Analysis.} The model successfully avoids commas in the creative-writing subtask (rewriting the sentence) but introduces two commas in the arithmetic subtask. This pattern, which we term \emph{domain-shift forgetting}, shows the constraint being maintained during the original IFEval task but dropped when the model transitions to mathematical reasoning. The math answer (33) is correct.

\subsection{Spatial Compartmentalization Under TriviaQA}
\label{app:ex_quotation}

DeepSeek-V3.1, natural-embedding condition with TriviaQA (C2-nat). The constraint is to wrap the \emph{entire} response in double quotation marks.

\begin{examplebox}[Example: Spatial Compartmentalization]
\small
\colorbox{blue!6}{\parbox{\dimexpr\linewidth-2\fboxsep}{%
\textbf{Prompt}\\[2pt]
\texttt{Write a speech-like paragraph on the US elections. Make sure to wrap your entire response with double quotation marks.}\\
\texttt{Additionally, please also complete the following:}\\
\texttt{`Boat Race' is Cockney rhyming slang for which part of the body?}
}}

\vspace{4pt}
\colorbox{red!6}{\parbox{\dimexpr\linewidth-2\fboxsep}{%
\textbf{Response}\\[2pt]
\texttt{"Fellow citizens, we stand at a pivotal moment in our nation's journey, a time when the very future of our democracy is placed in our hands. [\ldots] a shared commitment to the enduring promise of America."}\\
\texttt{"Boat Race" is Cockney rhyming slang for `face'.}
}}
\end{examplebox}

\paragraph{Analysis.} The model correctly wraps the speech paragraph in quotation marks, then produces a \emph{separate} line for the trivia answer (also with structural quotation marks around ``Boat Race''). Because the two parts are not enclosed in a single pair of quotation marks, the entire response fails the strict checker. This illustrates \emph{spatial compartmentalization}: the model treats the dual task as two independent outputs, applying the constraint locally rather than globally.

\subsection{Structural Abandon Under Arithmetic}
\label{app:ex_abandon}

o4-mini, natural-embedding condition with GSM8K (C4-1-nat). The constraint requires exactly 3 markdown bullet points using the \texttt{*}~marker.

\paragraph{Analysis.} The model correctly produces three \texttt{*}-marked bullet points for the naming subtask, matching the required format. However, the arithmetic breakdown uses a different list format (\texttt{-}~dashes), adding four dash-prefixed items. The IFEval checker counts all markdown list markers, yielding seven list items rather than the required three, a format violation introduced entirely by the distractor task's content. The math answer (34) is correct.

\begin{examplebox}[Example: Structural Abandon]
\small
\colorbox{blue!6}{\parbox{\dimexpr\linewidth-2\fboxsep}{%
\textbf{Prompt}\\[2pt]
\texttt{Name exactly 3 names for a black and white dog using markdown bullet points such as:}\\
\texttt{* Bullet point 1}\\[4pt]
\texttt{Additionally, please also complete the following:}\\[4pt]
\texttt{As Sally walked to school, she was holding the strings to 25 red balloons, 7 green balloons, and 12 yellow balloons. Suddenly, a gust of wind caused 40\% of the red balloons to burst. [\ldots] What number of balloons did she finally carry into the school?}
}}

\vspace{4pt}
\colorbox{red!6}{\parbox{\dimexpr\linewidth-2\fboxsep}{%
\textbf{Response}\\[2pt]
\texttt{* Oreo}\\
\texttt{* Panda}\\
\texttt{* Domino}\\[4pt]
\texttt{Sally's balloon count step by step:}\\
\texttt{- Red balloons: 25 $-$ (40\% of 25 = 10) = 15}\\
\texttt{- Green balloons: 7}\\
\texttt{- Yellow balloons: 12 $-$ (1/2 of 12 = 6) = 6}\\
\texttt{- Blue balloons added: 75\% of 8 = 6}\\[4pt]
\texttt{Total carried into school = 15 + 7 + 6 + 6 = \textbf{34} balloons.}
}}
\end{examplebox}


\subsection{Terminal Omission Under CNN/DailyMail}
\label{app:ex_terminal}

o4-mini, natural-embedding condition with CNN/DailyMail (C5-nat). The constraint requires the response to end with ``Peace!'' with no additional words following.

\paragraph{Analysis.} The model produces a complete rap that ends precisely with ``Peace!'' at the natural artistic conclusion. However, rather than stopping, it generates a five-bullet CNN/DailyMail summary (135 additional tokens follow ``Peace!''). The terminal constraint is satisfied \emph{within the first subtask} but violated \emph{globally}. This pattern (which we term \emph{terminal omission}) is especially prevalent under long-context summarization (C5-nat), where substantial additional content creates strong generation pressure past the intended stopping point. The failure is consistent across all three experimental runs with this IFEval item (ROUGE-L~=~0.18), demonstrating the robustness of this failure mode.

\begin{examplebox}[Example: Terminal Omission]
\small
\colorbox{blue!6}{\parbox{\dimexpr\linewidth-2\fboxsep}{%
\textbf{Prompt}\\[2pt]
\texttt{Write a strange rap song about Alexander the Great becoming the king of Macedon. Finish the song with:}\\
\texttt{Peace!}\\
\texttt{No additional words should follow ``Peace!''}\\[4pt]
\texttt{Additionally, please also complete the following:}\\[4pt]
\texttt{Please read the following article and write a concise summary of the key points. Article: Gordon Robson, 26, killed grandfather John Potts, 45, with a single punch [\ldots]}
}}

\vspace{4pt}
\colorbox{red!6}{\parbox{\dimexpr\linewidth-2\fboxsep}{%
\textbf{Response}\\[2pt]
\texttt{[\ldots six rap verses\ldots]}\\
\texttt{Kings and ghosts bow low to the madness of my throne--}\\
\texttt{I'm Alexander the Great, in every cosmic bone!}\\
\texttt{\textbf{Peace!}}\\[4pt]
\texttt{Summary of Key Points:}\\
\texttt{- Gordon Robson, 26, killed John Potts, 45, with a single punch [\ldots]}\\
\texttt{- Robson had been drinking after his grandfather's funeral [\ldots]}\\
\texttt{- Potts, a father-of-six, stepped in to calm the fight [\ldots]}\\
\texttt{- Robson admitted manslaughter; sentenced to three and a half years.}
}}
\end{examplebox}


\end{document}